\documentclass[12pt]{article}

\usepackage{sbc-template}

\usepackage{graphicx,url}

\usepackage[utf8]{inputenc}

\usepackage{tabularx}
\usepackage{xcolor}

\usepackage[colorlinks]{hyperref}
\hypersetup{
    colorlinks=false,
    allbordercolors={white},
}

\definecolor{light-gray}{gray}{0.95}

\newcommand{\F}{\textit{F}\textsubscript{1}}

\author{Diogo Cortiz\inst{1,2}, Jefferson O. Silva\inst{2,4}, Newton Calegari\inst{2}, Ana Luísa Freitas\inst{3}, \\

Ana Angélica Soares\inst{3}, Carolina Botelho\inst{3}, Gabriel Gaudencio Rêgo\inst{3}, \\

Waldir Sampaio\inst{3}, Paulo Sergio Boggio\inst{3}}

\address{Brazilian Network Information Center (NIC.br) \\
  São Paulo, SP -- Brazil
  \email{diogo@nic.br}
\nextinstitute
  Pontifical Catholic University of São Paulo (PUC-SP) \\
  São Paulo, SP -- Brazil
  \email{\{dcortiz, silvajo, njcalegari\}@pucsp.br}
\nextinstitute
  Mackenzie Presbyterian University \\
  São Paulo, SP -- Brazil
  \email{\{paulo.boggio\}@mackenzie.br}
\nextinstitute
    Jusbrasil \\
    São Paulo, SP -- Brazil
}

\title{A Weakly Supervised Dataset of Fine-Grained Emotions in Portuguese}

\begin{document}

\maketitle

\begin{abstract}
Affective Computing is the study of how computers can recognize, interpret and simulate human affects. Sentiment Analysis is a common task in NLP related to this topic, but it focuses only on emotion valence (positive, negative, neutral). An emerging approach in NLP is Emotion Recognition, which relies on fined-grained classification. This research describes an approach to create a lexical-based weakly supervised corpus for fine-grained emotion in Portuguese. We evaluate our dataset by fine-tuning a transformer-based language model (BERT) and validating it on a Gold Standard annotated validation set. Our results (F1-score= $.64$) suggest lexical-based weak supervision as an appropriate strategy for initial work in low resourced environment.
\end{abstract}

\begin{resumo}
A Computação Afetiva é o estudo de como os computadores podem reconhecer, interpretar e simular os afetos humanos. A Análise de Sentimento é uma tarefa comum em PLN, mas se concentra apenas na valência da emoção (positiva, negativa, neutra). Uma abordagem emergente é o Reconhecimento de Emoção, que depende de uma classificação refinada. Nesta pesquisa, descrevemos uma abordagem de supervisão fraca baseada em Itens Lexicais para criar um corpus de emoções refinadas em português. Avaliamos nosso corpus fazendo o ajuste fino de um modelo de linguagem baseado em Transformer (BERT) e avaliando-o em um conjunto de validação anotado. Nossos resultados (F1-score= $.64$) sugerem que a supervisão fraca baseada em Itens Lexicais pode ser uma estratégia apropriada para o trabalho inicial em ambiente de poucos recursos.

\end{resumo}

\section{Introduction}
\label{sec:introduction}

Affective Computing comprises the study of how computers can recognize, interpret and simulate human affects. According to Picard \cite{Rosalind2000}, a field pioneer, it is imperative to develop ways for computers to be able to recognize, understand and express emotions for intelligent and natural interaction between humans and machines. Although Affective Computing may employ several input types such as facial expression images, voice, or physiological data, our research focuses on the written language. Thus, our scope lies in the area of Natural Language Processing (NLP).

A common task in NLP is Sentiment Analysis which classifies a text into three different categories: positive, negative, and neutral \cite{Drus2019}. The Emotion Recognition task, however, is a more detailed NLP task. Rather than classifying text only into valence categories (positive, negative, or neutral), it classifies text into more detailed emotional categories.

The delimitation of this research is concentrated in the area of fine-grained Emotion Recognition. We studied an approach to create a corpus in Portuguese for this task using a weak supervision approach.

\section{Related Work}
\label{sec:related-work}

Several works in NLP are based on the theory of basic emotions \cite{Ekman1992} to classify texts into defined categories of emotions, the number of categories ranging from 4 (four) to 8 (eight). One of the studies adopted the 6 (six) basic emotions proposed by Ekman (joy, fear, anger, sadness, surprise, and disgust) to train an Emotion Recognition model \cite{Batbaatar2019}. Another research added trust and anticipation to the basic emotions, working with 8 (eight) basic emotion categories [Sosea and Caragea 2020]. 

In the realm of emotion study, other relevant theory is the Theory of
Constructed Emotion \cite{Barrett2016}, which assumes that emotions are not universal, but idiosyncratic. This theoretical debate imposes methodological limitations that a
different computational approach may help to solve. The Semantic Space Theory
\cite{Cowen2021} let us recognize and analyze emotional content of naturalistic stimuli, using open-ended statistical techniques to capture emotional variations in behavior. The results suggest that more than 25 emotional classes have distinct profiles of previous expressions and events. The authors argue that these emotions are high-dimensional, categorical, and often blended.

GoEmotions is a dataset with more than 58,000 English Reddit comments for training NLP models in the Emotion Recognition task. It was annotated for 27 categories of emotions and Neutral based on the Semantic Space Theory. The authors fine-tuned a BERT language model and achieved an average \F-score of $.46$ \cite{Demszky2020}.

Despite the average \F-score below $.50$, some classes scored above $.70$. It is a complex dataset with many categories that often have fuzzy boundaries between them. It is essential to discuss the importance of creating a fine-grained dataset with more emotional categories. Research in the Affective Computing area is not limited to Sentiment Analysis or categories proposed by the basic theory.

Although the GoEmotion dataset was released according to open data standards \cite{Demszky2020}, the scope of the corpus is limited to English, which makes it difficult to use in applications in other languages. One of the challenges that Machine Learning faces is dealing with a low-resourced environment (when the data available is not enough to train the models). This phenomenon can happen in specific domains of applications but also in specific geographic regions.

In the area of NLP, there is a lack of datasets and corpus available in many languages. It is the case of Portuguese, which has a small amount of Sentiment Analysis datasets when compared to English \cite{Pereira2021}. It is worth noting that we searched for a dataset of fine-grained emotion, but we did not find any in Portuguese.

\section{Objectives, Research Questions and Hypothesis}
\label{sec:objectives}

This research aims to study the creation of a corpus of fine-grained emotions for low resourced languages, specifically Portuguese. Due to limited financial resources, a specific objective of this work is to study the use of the weak supervision strategy to construct our corpus. Weak supervision is a strategy when there is no human annotation of each data point, but the labels are attributed using noisy and limited sources or specific rules. We proposed the following research questions (RQ) to guide our work:

\noindent
RQ1: Is the weak supervision strategy suitable for building an NLP corpus for the fine-grained Emotion Recognition task in a low resourced environment?

\noindent
RQ2: What is a proper weak supervision approach to construct a corpus for fine-grained Emotion Recognition tasks in NLP?

Our first hypotheses (H1) is that weak supervision could be a suitable strategy to build NLP corpus for emotion recognition. Our second hypotheses (H2) is that lexical-based approach can be an adequate strategy to collect samples for each of the categories of our dataset, using the Lexical Items (LI) as a criterion for defining the label in an adequate way for Portuguese. A third hypothesis (H3) is that using SOTA Machine Learning techniques (specifically Transformers-based language models), combined with masking techniques in the LI presented in the weakly supervised corpus, can avoid the model overfit the learning phase.

To answer RQ1 and RQ2 and validate our hypotheses, we prepared an experiment to create a weakly supervised corpus in Portuguese and measure its performance by training a classification model. The following sections will describe our experimental protocol, including how we collected and weakly annotated the data, our model architecture, metrics, and results.

\section{Experimental Protocol}
\label{sec:experimental-protocol}

Our experiment is composed of the following pipeline: defining emotion categories based on semantic space theory for Portuguese; selecting the lexical items related to each emotion category based on its definition; collecting the data; manually annotating a test dataset to create a gold standard; defining the model architecture; training the model and evaluating it on the gold standard. Each of them is described in detail in this paper.

\subsection{Defining Emotion Categories}

The emotion categories for this research were defined from a review of the GoEmotion work \cite{Demszky2020}. The review process had two stages and the participation of a group of 7 (seven) researchers with different backgrounds (psychology, neuroscience, sociology, communications, cognitive science, and computer science). In the first stage,  the researchers discussed and reviewed each emotion in English during a working meeting. They proposed a translation into Portuguese based on the definitions of each emotion. The result of this first stage was a translated list of terms with consensus among the reviewers.

The second stage was reviewing the categories' definitions in Portuguese to check if they were consistent with the language. The reviewers suggested changing the emotional category \textit{cuidado}, translated from \textit{caring} to \textit{compaixão}, as it is a more broad and blended category in the Portuguese language. The second proposal was the removal of the emotion \textit{realization}, in the sense of perceiving something, as it is not a much prevalent emotional category in the Portuguese language. Finally, there was a consensus among researchers to add the categories \textit{saudade} and \textit{inveja} to the list. We also removed neutral to focus on emotions. The final list consists of 28 emotional categories in total. All emotions and their definitions are presented in Table \ref{tab:emotion-categories}.

\begin{table}[ht]
  \caption{Portuguese Emotion Categories}
  \label{tab:emotion-categories}
  \begin{tabularx}{\textwidth}{X}
    \centering
    \includegraphics[trim=1.2cm 10.7cm 2.2cm 1.5cm, clip, scale=0.7]{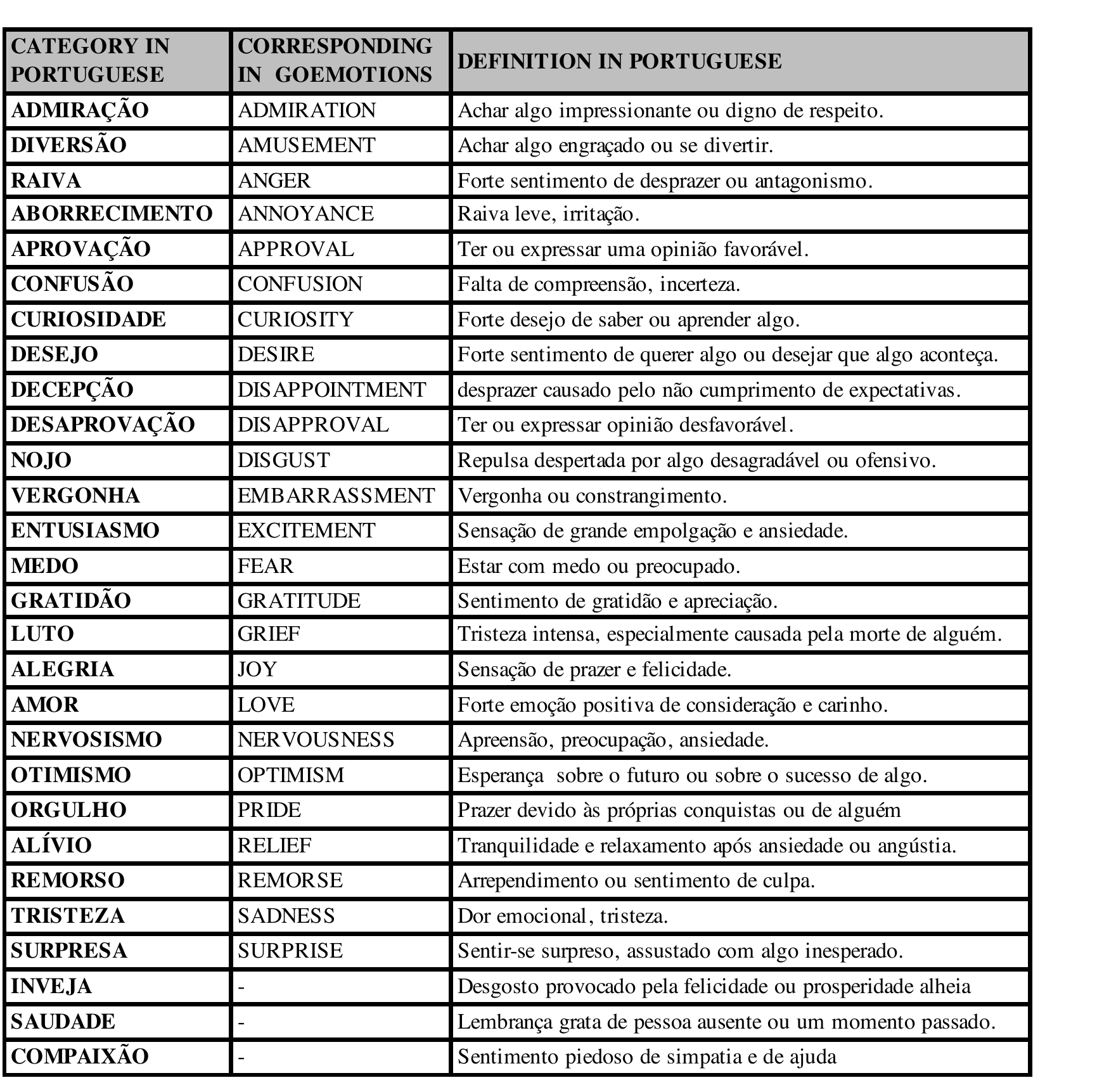}
  \end{tabularx}
\end{table}

\subsection{Selecting Lexical Items for weak supervision }
After translating and defining emotions into Portuguese, the next step was to select the Lexical Items that would serve as a filter to search for examples and label assignment rules (weak supervision). For each of the emotions on the list, we initially look for related Lexical Items by synonyms. For this, we use the database available at www.sinonimos.com.br, which has more than 30 thousand synonyms of words and expressions for Portuguese.

Because some words of emotion presented polysemic behavior, we opted for human curation to select the proper Lexical Items. Only synonyms with a semantic relationship with the definition of emotion were considered. For each Verbal Lexical Item, we collect the different conjugations in the repository www.conjugacao.com.br to cover all tenses and moods in Portuguese. To avoid the negation effect, we manipulated the data as follows: we searched for the combination of the word "não" (no/not in Portuguese) or "nem" (neither in Portuguese) followed by a Lexical Item in our list. If an example was found, we removed it from our dataset. We also added slang and terms related to emotions that were known to the authors. The result of this step was a list in which each emotion was associated with a set of lexical items, which were later used as a data collection filter and label assignment rule.

\subsection{Data}

We use Twitter as a data source. The collection was made between the 23rd and 24th of June (2021) using the platform's official API. The filters used were the list of terms associated with each emotion. Retweets and replies were not considered, keeping only original tweets. Hashtags were removed, but emojis were kept.

In total, 49179 tweets were collected using a weak supervision approach. Each example received the category label according to the Lexical Item used in the collection. For example, if a tweet was collected because it was filtered by a term associated with the emotion \textit{amor}, it would be labeled to the \textit{amor} category.

\begin{figure}[ht]
  \label{fig:example-per-category}
  \caption{Examples per categories.}
  \includegraphics[scale=0.7]{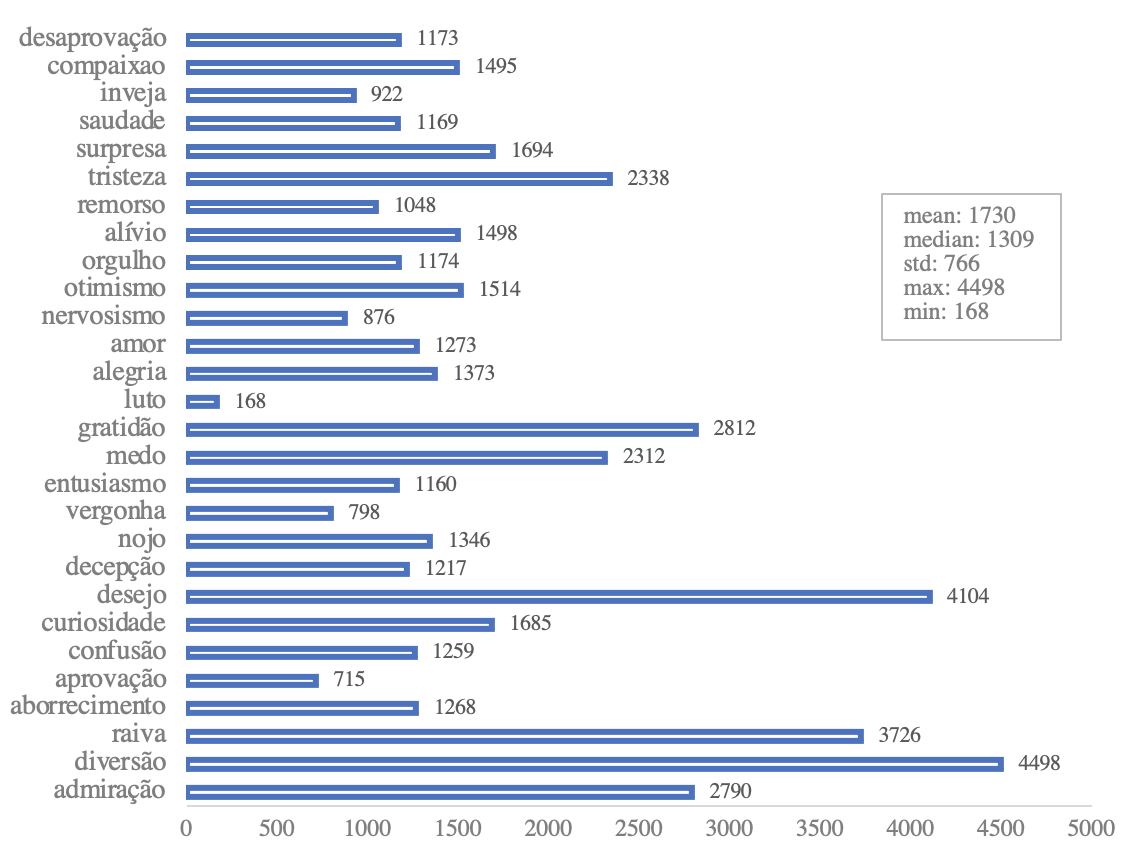}
  \centering
\end{figure}

We tried to maintain a balanced distribution of examples among the classes, but the results of our collection process suggest that some emotional categories are more prevalent than others. We intend to focus on additional data collection for the categories with the smallest number of examples to achieve a better balance distribution in future work. For the training set, we had a total of 47405 examples. We present in Figure \ref{fig:example-per-category} the total number of examples by category and the descriptive statistics of our dataset \footnote{Data available at: https://github.com/diogocortiz/PortugueseEmotionRecognitionWeakSupervision }.

\subsubsection{Masking Lexical Itens}
A hypothesis that appeared during the execution of this research was that the models could memorize the Lexical Items (LI) associated with each emotion, reducing generalization properties and causing the model to overfit. We chose to apply a masking technique to the Lexical Items used for collection and label assignment to investigate this phenomenon. The masking technique consisted of replacing an LI by [MASK], as can be seen in the examples in Table 2.

\begin{table}[ht]
  \label{tab:example-masked}
  \caption{Masking technique}
  \begin{tabular}{|l|l|}
    \hline
    \textbf{Original} & tô indignada e não é pouco! \\ \hline
    \textbf{Masked} & tô {[}MASK{]} e não é pouco! \\ \hline
  \end{tabular}
  \centering
\end{table}

We ended up with three datasets for training three different models. The first is the original dataset that we created using the weakly supervised approach without any masking technique. We identified this dataset as NoMask. The second dataset is the result of applying the masking technique to 30\% of examples for each category. We identified this dataset as 30Mask. The third dataset is the result of masking all Lexical Items. We identified this dataset as FullMasked.

\subsubsection{Gold standard for validation}
Despite this research studying the feasibility of the weak supervision approach for the Emotion Recognition task in NLP, it is worth noting the importance of building a dataset with human curation to evaluate the performance of a trained model with the created dataset.

To meet this requirement, we separated a set composed of 1773 examples from the dataset created earlier; we removed the labels assigned by the weak supervision approach so that a human could manually annotate them. We did not apply any Masking technique to this set. Due to limited resources, it was not possible to cross-annotate the validation dataset. Only one annotator annotated each example. For this reason, it is not possible to present any measure of agreement between the annotators. We recognize the limitations of this procedure, which can reduce the quality of supervision and introduce bias.

\subsection{Models}

To study the performance of our dataset, we needed to fine-tune the BERT language model to the Emotion Recognition task using our weakly supervised dataset. The Bidirectional Encoder Representations from Transformers (BERT) pre-trained language model \cite{Devlin2018} was released by Google in 2018. Since then, the use of this architecture has improved the performance in different natural language tasks. In our research, we used BERTimbau \cite{Bertimbau}, a pre-trained BERT model for Brazilian Portuguese. We fine-tuned three different models using our three different datasets (NoMask, 30Mask, and FullMask).

\subsubsection{Parameters Settings}
When finetuning the BERT language model, we keep most of the hyperparameters set in the original paper \cite{Devlin2018}. We changed only batch size and learning rate as proposed by \cite{Demszky2020}. We trained each model for 4 (four) epochs. The threshold to set a classification as positive was $.30$ (the same used by \cite{Demszky2020}). All models were implemented using the huggingface library. The training process used the same computing environment (Quadro RTX 6000).

\subsection{Results}

The results in Table 3 show the performance of our three models. As we can observe, the model trained with 30\% of LI masked (30Mask) had a similar performance to the model trained with the original dataset with no intervention (NoMask).

\begin{table}[ht]
  \label{tab:results-weak-supervision}
  \caption{Results based on weak supervision}
  \begin{tabularx}{\textwidth}{X}
    \centering
    \includegraphics[trim=0.2cm 9.8cm 9.6cm 1.5cm, clip, scale=0.7]{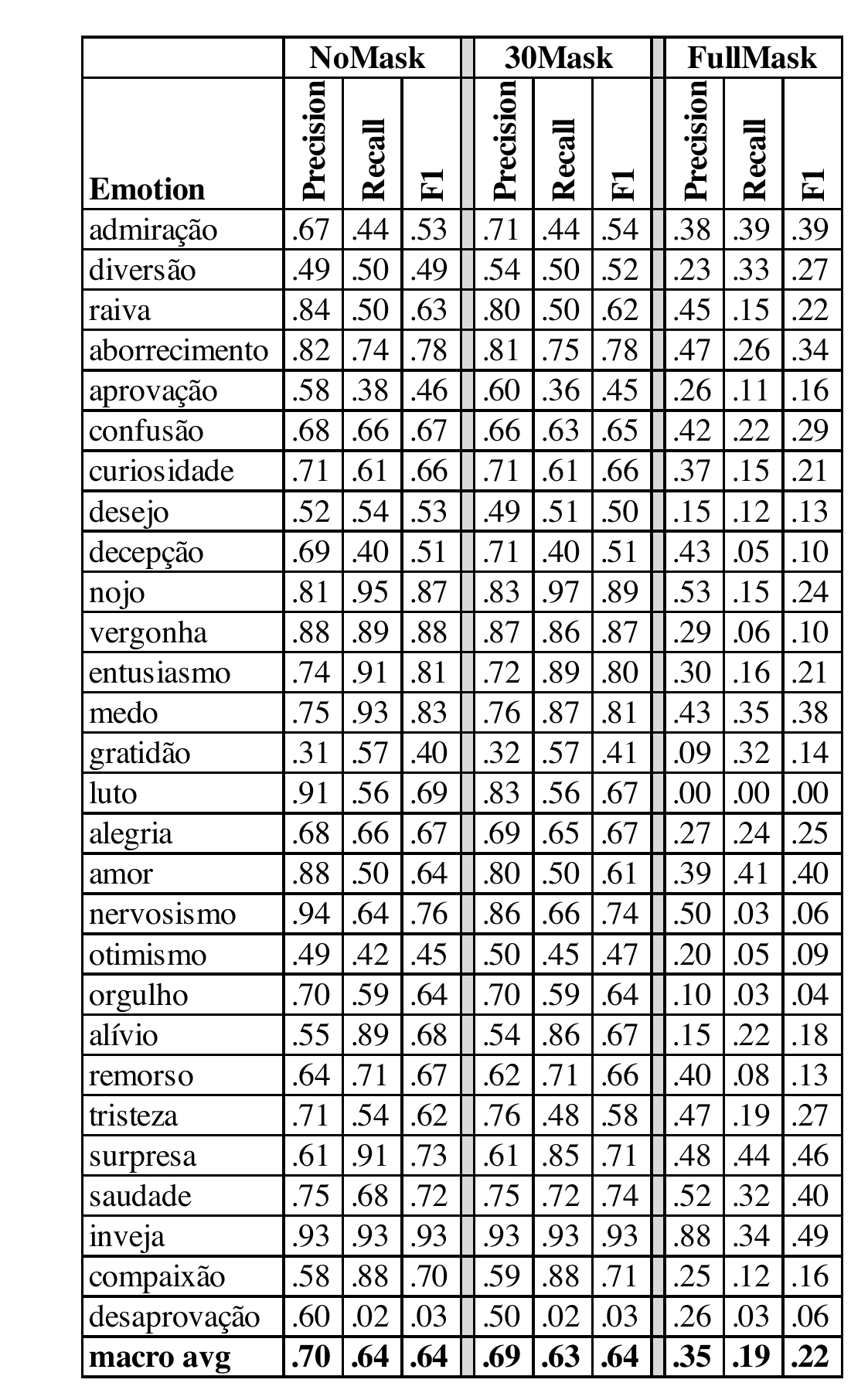}
  \end{tabularx}
\end{table}

Those results suggest that masking a certain amount of Lexical items used as a label rule (weak supervision) could be an appropriate strategy to stimulate the model to learn by context and not only by memorizing LI. However, masking all LI introduces much noise to the training dataset, significantly impacting the model performance.

\section{Conclusions}

According to the results presented, we argue that the adoption of Weak Supervision may be an appropriate strategy for some NLP activities in low-resource scenarios. The creation of datasets is costly and often prohibitive for some economies, making weak supervision an initial alternative for projects when there are insufficient resources to adopt a human supervision methodology. Our RQ1 inquired whether weak supervision is a proper approach to construct a corpus for fined-grained Emotion Recognition in low resourced environment. We found consistent results when evaluating our models, suggesting that weak supervision is an appropriate approach for initial work in the Emotion Recognition NLP task in Portuguese. The results supports our first hypotheses (H1).

This research used a Lexical-based approach to collect, and weak supervise the dataset. According to the results achieved and based on our empirical experience during its execution, we argue that this approach can be appropriate for collecting and annotating data in tasks involving narrow scenarios and well-defined problems. The results help us to answer our RQ2 and validate the H2. However, our experiment has some limitations, such as the validation dataset created from the initial collection of Lexical Items, which makes it difficult to assess the generalization performance of the models. In this sense, we can neither validate nor refute our third hypothesis (H3). We plan to build a new dataset with human supervision in future work without using the Lexical Items list in the filter during data collection. It will be possible to validate and compare the generalization performance of models using different datasets.

\bibliographystyle{sbc}
\bibliography{referencias}

\end{document}